\documentclass{article}

\usepackage[final,nonatbib]{neurips_2023}

\usepackage[utf8]{inputenc} 
\usepackage[T1]{fontenc}    
\usepackage{hyperref}       
\usepackage{url}            
\usepackage{booktabs}       
\usepackage{amsfonts}       
\usepackage{nicefrac}       
\usepackage{microtype}      
\usepackage{xcolor}         
\usepackage{color, colortbl}

\usepackage{listings}
\usepackage{xcolor}

\definecolor{LightCyan}{rgb}{0.88,1,1} 
\definecolor{Pink}{rgb}{1, 0.71, 0.756} 
\definecolor{PaleGreen}{rgb}{0.8, 1, 0.8} 
\definecolor{PaleBlue}{rgb}{0.8, 0.9, 1} 
\definecolor{Orange}{RGB}{255, 180, 130} 
\definecolor{Purple}{RGB}{208, 187, 255} 
\definecolor{BrightPink}{RGB}{250, 176, 228} 
\definecolor{Yellow}{RGB}{255, 254, 163} 
\definecolor{Brown}{RGB}{222, 187, 155} 

\usepackage{times}
\usepackage{epsfig}
\usepackage{graphicx}
\usepackage{amsmath}
\usepackage{amssymb}
\usepackage{diagbox}
\usepackage{adjustbox}
\usepackage{xltabular}

\usepackage[symbol]{footmisc}

\usepackage{tikz}
\usepackage{caption}
\usepackage{subcaption}

\usepackage{wrapfig}
\usepackage{comment}

\usepackage[textsize=tiny]{todonotes}
\usepackage{blindtext}
\usepackage{multirow}
\usepackage{makecell}

\definecolor{light-gray}{gray}{0.95}

\usepackage{graphicx}
\usepackage{array}
\usepackage{booktabs}
\newcolumntype{N}{>{\centering\arraybackslash}m{.5in}}
\newcolumntype{G}{>{\centering\arraybackslash}m{1in}}
\newcolumntype{H}{>{\centering\arraybackslash}m{1.1in}}
\newcolumntype{M}{>{\centering\arraybackslash}m{.3in}}
\newcolumntype{P}{>{\centering\arraybackslash}m{.8in}}
\newcolumntype{Q}{>{\centering\arraybackslash}m{.4in}}

\begin{document}

\title{Stable Code Technical Report}

\author{
Nikhil Pinnaparaju \quad Reshinth Adithyan \quad Duy Phung \quad Jonathan Tow\\
James Baicoianu \quad Ashish Datta \quad  Maksym Zhuravinskyi   \\
Dakota Mahan \quad Marco Bellagente \quad Carlos Riquelme \quad Nathan Cooper
\\
\\
\textbf{Stability AI Language Team}\thanks{Correspondance to: \{nikhil.pinnaparaju, reshinth, nathan.cooper\}@stability.ai}
}
\maketitle

\begin{abstract}
We introduce Stable Code, the first in our new-generation of code language models series, which serves as a general-purpose base code language model targeting code completion, reasoning, math, and other software engineering-based tasks. Additionally, we introduce an instruction variant named Stable Code Instruct that allows conversing with the model in a natural chat interface for performing question-answering and instruction-based tasks.
In this technical report, we detail the data and training procedure leading to both models.
Their weights are available via Hugging Face for anyone to download and use \footnote{\url{https://huggingface.co/stabilityai/stable-code-3b}}\footnote{\url{https://huggingface.co/stabilityai/stable-code-instruct-3b}}.
This report contains thorough evaluations of the models, including multilingual programming benchmarks, and the MT benchmark focusing on multi-turn dialogues.
At the time of its release, Stable Code is the state-of-the-art open model under 3B parameters and even performs comparably to larger models of sizes 7 billion and 15 billion parameters on the popular Multi-PL benchmark. Stable Code Instruct also exhibits state-of-the-art performance on the MT-Bench coding tasks and on Multi-PL completion compared to other instruction tuned models.
Given its appealing small size, we also provide throughput measurements on a number of edge devices.
In addition, we open source several quantized checkpoints and provide their performance metrics compared to the original model.
\end{abstract}
\section{Introduction}
The application of Machine Learning in programming languages has been prolific: from code understanding to code representation or code completion.
Earlier work focused on exploiting the underlying deep semantic structure of programming languages like in Code2Vec \cite{Alon2018code2vecLD}, Code2Seq \cite{Alon2018code2seqGS}, Graph Representation Learning for Code \cite{Allamanis2017LearningTR}.
The above architectures are tailor-made for the native structures of Abstract Syntax Trees (AST) / Data Flow Graphs (DFG), and have a significant limitation: they can only be applied for tasks that involve completely executable code.
For instance, to generate DFG as outlined by Allamanis et al.\cite{Allamanis2017LearningTR}, the program needs to be complete. Works such as Austin et al.,~\cite{austin2021program} have shown how transformers-based models can be used like natural language for code at the lexical (text) level. Since then, language models have been widely used to model code on a variety of tasks.
The extensible training dynamics of language models offer improved reasoning and understanding ability \cite{yang2024wand} often captured via their application in productivity tools for software engineers such as code completion \cite{githubcopilot, hou2023large}, question answering \cite{cursor, githubcopilot}, and debugging plugins \cite{chen2023teaching, hou2023large}. Models like these are executed every few seconds, especially in the case of code completion.
Accordingly, strong models that can run on consumer devices are preferred, as avoiding network latency makes a difference and addresses various discrepancies with respect to gated APIs.
In this work, we introduce Stable Code and Stable Code Instruct, small and fast code language models that achieves strong performance on popular benchmarks while matching significantly larger models.

The report is structured as follows.
Section \ref{sec:pt-data} describes the training data in detail, and the architectural choices are provided in Section \ref{sec:arc}.
Section \ref{sec:training-base} covers the training procedure applied to the base Stable Code model, while Section \ref{sec:alignment} focuses on instruction tuning.
In Section \ref{sec:results} we share the performance of these models and Section \ref{sec:inference} is centered around throughput and quantization.
Finally, Section \ref{sec:conclusion} concludes.

\begin{table}[ht]
\centering
\noindent
\begin{tabular}{G N G N N N}
\toprule
\multicolumn{1}{ l }{\textbf{Dataset}} & \textbf{Sampling Weight} & \textbf{Num Tokens Sampled} & \textbf{Epochs} & \textbf{Category}\\[1ex] 
\midrule
\rowcolor{Yellow}
\multicolumn{1}{ l }{\textbf{StarCoder C}} & 0.0924 & 122,202,657,912.00 &6.0& Code\\ [0.2ex]
\rowcolor{Yellow}
\multicolumn{1}{ l }{\textbf{StarCoder CPP}} & 0.0734 & 97,032,316,152.00 &6.0& Code\\ [0.2ex] 
\rowcolor{Yellow}
\multicolumn{1}{ l }{\textbf{StarCoder Java}} & 0.1029 & 136,010,698,326.00 &6.0& Code\\ [0.2ex]
\rowcolor{Yellow}
\multicolumn{1}{ l }{\textbf{StarCoder Javascript}} & 0.0858 & 113,469,977,934.00 &6.0& Code\\ [0.2ex]
\rowcolor{Yellow}
\multicolumn{1}{ l }{\textbf{StarCoder CSS}} & 0.0146 & 19,285,266,328.00 &4.0& Code\\ [0.2ex]
\rowcolor{Yellow}
\multicolumn{1}{ l }{\textbf{StarCoder Go}} & 0.0258 & 34,092,166,492.00 &4.0& Code\\ [0.2ex]
\rowcolor{Yellow}
\multicolumn{1}{ l }{\textbf{StarCoder HTML}} & 0.0298 & 39,354,336,188.00 &4.0& Code\\ [0.2ex]
\rowcolor{Yellow}
\multicolumn{1}{ l }{\textbf{StarCoder Ruby}} & 0.0061 & 8,011,730,332.00 &4.0& Code\\ [0.2ex]
\rowcolor{Yellow}
\multicolumn{1}{ l }{\textbf{StarCoder Rust}} & 0.0122 & 16,131,445,656.00 &6.0& Code\\ [0.2ex]
\rowcolor{Yellow}
\multicolumn{1}{ l }{\textbf{StarCoder Markdown}} & 0.1154 & 152,629,435,716.00 &6.0& Code\\ [0.2ex]
\rowcolor{Yellow}
\multicolumn{1}{ l }{\textbf{StarCoder Shell}} & 0.0033 & 4,323,112,416.00 &4.0& Code\\ [0.2ex]
\rowcolor{Yellow}
\multicolumn{1}{ l }{\textbf{StarCoder Php}} & 0.0764 & 100,958,420,706.00 &6.0& Code\\ [0.2ex]
\rowcolor{Yellow}
\multicolumn{1}{ l }{\textbf{StarCoder Sql}} & 0.0247 & 32,645,285,202.00 &6.0& Code\\ [0.2ex]
\rowcolor{Yellow}
\multicolumn{1}{ l }{\textbf{StarCoder R}} & 0.0003 & 415,957,896.00 &4.0& Code\\ [0.2ex]
\rowcolor{Yellow}
\multicolumn{1}{ l }{\textbf{StarCoder Typescript}} & 0.0224 & 29,634,722,636.00 &4.0& Code\\ [0.2ex]
\rowcolor{Yellow}
\multicolumn{1}{ l }{\textbf{StarCoder Python}} & 0.1067 & 141,067,150,184.00 &8.0& Code\\ [0.2ex]
\rowcolor{Yellow}
\multicolumn{1}{ l }{\textbf{StarCoder Jupyter}} & 0.0060 & 7,941,540,044.00 &4.0& Code\\ [0.2ex]
\rowcolor{Yellow}
\multicolumn{1}{ l }{\textbf{StarCoder Restructured Text}} & 0.0032 & 4,179,202,492.00 &4.0& Code\\ [0.2ex]
\midrule

\rowcolor{Purple}
\multicolumn{1}{ l }{\textbf{Github Issues}} &0.0231&46,302,993,820 &2.5& Technical\\ [0.2ex]
\rowcolor{Purple}
\multicolumn{1}{ l }{\textbf{Github Diffs}} &0.0019&3,817,060,582 &2.0& Technical\\ [0.2ex] 
\rowcolor{Purple}
\multicolumn{1}{ l }{\textbf{StackExchange}} &0.0019&3,817,060,582 &2.0& Technical\\ [0.2ex] 
\midrule
\rowcolor{Brown}
\multicolumn{1}{ l }{\textbf{Synthetic}} & 0.0006 & 819,864,748.00 &3.0& Technical\\ [0.2ex] 
\midrule

\rowcolor{LightCyan}
\multicolumn{1}{ l }{\textbf{Proof Pile}} & 0.0384 & 50,780,637,096 &1.0& Math\\ [0.2ex] 
\rowcolor{LightCyan}
\multicolumn{1}{ l }{\textbf{Meta Math QA}} & 0.0003 & 83,663,501 &4.0& Math\\ [0.2ex] 
\midrule
\rowcolor{Pink}
\multicolumn{1}{ l }{\textbf{Arxiv}} & 0.0213 & 28,097,511,912.00 &1.0& Web\\ [0.2ex] 
\rowcolor{Pink}
\multicolumn{1}{ l }{\textbf{Refined Web}} & 0.0220 & 29,114,185,066.13 &0.5& Web\\ [0.2ex] 
\midrule
\multicolumn{1}{ l }{\textbf{Total}} & 1 &1,322,090,182,830.13 && -\\ [0.2ex] 
\end{tabular}
\caption{The complete Stable Code 3B training set with sampling weights. The tokens count refers to the \textbf{NeoX} tokenizer introduced in Sec.~\ref{sec:tokenizer}}
\label{tab:data-mix}
\end{table}

\section{Training Data}\label{sec:pt-data}
\subsection{Pretraining Dataset}
To construct the pre-training dataset for Stable Code, we collected a diverse array of publicly accessible, large-scale data sources. These sources encompass a wide spectrum of code repositories, extensive collections of technical documents (example: readthedocs), mathematically focused texts, and comprehensive web datasets. The primary objective of this initial \textit{pretraining} phase is to learn a rich internal representation that extends beyond mere code comprehension. Our aim was to significantly enhance the model's capabilities in mathematical understanding, logical reasoning, and processing of complex technical texts surrounding software development. The rationale behind selecting such a varied dataset mix is to develop a language model well suited to perform a wide range of software engineering tasks, not only those directly related to programming like code completion.




Furthermore, our training data also contains general text datasets in order to provide the model with a broader linguistic knowledge and context.
We hope this enables the model to address a wider range of queries and tasks in a conversational manner.
In Table \ref{tab:data-mix}, we provide the data sources, epochs, categories, and sampling weights of the datasets used to set up the pretraining corpus. We use an 80:20 split distribution of code and natural language data, respectively, with the contributions from individual components detailed in the table (see Table \ref{tab:data_refs} for references).

\begin{xltabular}{\textwidth}{l|c}
\caption{References for main training datasets.} \label{tab:data_refs} \\
\toprule
\textbf{Dataset} & \textbf{Reference} \\
\midrule
\endfirsthead
\toprule
\textbf{Dataset} & \textbf{Reference} \\
\endhead

\bottomrule
\multicolumn{2}{r}{{Continued on next page}} \\
\endfoot

\bottomrule
\endlastfoot

StarCoder Data & ~\cite{li:starcoder} \\
Github Issues & ~\cite{Kocetkov2022TheStack} \\
Github Diffs & ~\cite{muennighoff2023octopack} \\
Stackexchange & ~\cite{together2023redpajama} \\
Arxiv & ~\cite{together2023redpajama} \\
Synthetic Dataset & Sec ~\ref{sec:synthetic-pt} \\
Proof Pile Math & ~\cite{azerbayev2023llemma} \\
Meta Math QA & ~\cite{yu2023metamath} \\
Refined Web & ~\cite{penedo2023refinedweb} \\
\end{xltabular}



\subsubsection{Synthetic Dataset}\label{sec:synthetic-pt}
We also introduce a small synthetic dataset into our pre-training corpus. The data is synthetically generated from the seed prompts of the CodeAlpaca\footnote{Subset found here - https://huggingface.co/datasets/HuggingFaceH4/CodeAlpaca\_20K} dataset, which is comprised of 174,000 prompts. To augment the diversity and difficulty presented in the code-alpaca prompts, we employed the "Evol-Instruct" method as introduced by Xu et al.,~\cite{xu2023wizardlm} wherein we ask a language model (in this case, we use WizardLM ~\cite{xu2023wizardlm}) to increase the complexity of the given seed prompt in a step-by-step fashion. By applying strategies focused on breadth, reasoning, deepening, and complexity, we were able to enrich our collection with an additional 100,000 prompts. We leverage the DeepSeek Coder 34B model~\cite{guo2024deepseekcoder} to generate synthetic outputs for the newly developed "Evol-Instruct" prompts. We believe that introducing this synthetic data early during the pretraining phase helped the model respond better to natural language text based on an ablation experiment we conducted.

\subsection{Long Context Dataset}
Building upon the initial pre-training phase, we composed an additional stage of training that specifically targets the model's capability to process and comprehend long sequences. Having a longer context length is useful for coding models due to the usual inter-dependence of multiple files within a repository. 
We specifically chose 16,384 as the context length for our long context dataset after determining the median and mean number of tokens in a software repository to be $\approx12k$ and $\approx18k$ tokens, respectively.
This continued training stage focused on a curated selection of programming languages, all sourced from The Starcoder dataset~\cite{li:starcoder}, a filtered version of the Stack which is a large collection of high quality and permissively licensed coding data \cite{Kocetkov2022TheStack}. The languages selected for this phase were based on the Stack Overflow Developer Survey 2022 \cite{Stackoverflow}.
In particular, we selected \textit{Python, C, C++, Go, Java, and JavaScript}.

To create this long context dataset, we took files written in these languages within a repository and combined them, inserting a special \textit{<repo\_continuation>} token between each file to maintain separation while preserving the flow of content. To circumvent any potential biases that might arise from a fixed ordering of files—a factor that could inadvertently teach the model an unintended sequence or hierarchy—we employed a randomized strategy. For each repository, we generated not one, but two distinct orderings of the concatenated files.
The statistics are outlined in Table \ref{tab:ptdata-dist}.
\begin{xltabular}{\textwidth}{l|c|c}
\caption{Data distribution of files before and after concatenation} \label{tab:ptdata-dist} \\
\toprule
\textbf{Statistic} & \textbf{File Level} & \textbf{Long Context} \\
\midrule
\endfirsthead
\toprule
\rowcolor[HTML]{D9D9D9} 
\textbf{Statistic} & \textbf{File Level} & \textbf{Long Context} \\
\endhead

\bottomrule
\multicolumn{3}{r}{{Continued on next page}} \\
\endfoot

\bottomrule
\endlastfoot

Percentage of Rows \textgreater{}= 4096 & 5 & 100 \\
Median & 470 & 12834 \\
Max Length & 326833 & 1020069 \\
Min Length & 3 & 6008 \\
Mean & 1195 & 18000 \\
\end{xltabular}





\section{Model Architecture}\label{sec:arc}
Stable Code is built on top of Stable LM 3B \cite{StableLM-3B-4E1T}, which is a state-of-the-art LLM for natural language in English at the 3 billion parameter scale. 
The model is a causal decoder-only transformer similar in design to the LLaMA architecture~\cite{touvron2023llama}. Table~\ref{tab:architecture-layout} shows some of the key architectural details.
In particular, the main differences with respect to LLaMA are the following:
\begin{itemize}
    \item 
    \textbf{Position Embeddings}. Rotary Position Embeddings~\cite{su2023roformer} applied to the first $25\%$ of head embedding dimensions for improved throughput following~\cite{black2022gptneox20b}.
    \item 
    \textbf{Normalization}. LayerNorm~\cite{ba2016layer} with learned bias terms as opposed to RMSNorm~\cite{zhang2019root}.
    \item 
    \textbf{Biases}. We remove all bias terms from the feed-forward networks and multi-head self-attention layers, except for the biases of the key, query, and value projections~\cite{bai2023qwen}.
\end{itemize}

\subsection{Tokenizer}\label{sec:tokenizer}
We use the same tokenizer as the Stable LM 3B model, which is based on the BPE tokenizer from Black et al.,\cite{black2022gptneox20b} with a vocabulary size of 50,257. We also added the special tokens from the StarCoder models that include tokens for indicating the name of the file, the number of stars for the repository, Fill-in-Middle (FIM), etc. For our long context training stage, we added a special token to indicate when two concatenated files belong to the same repository.

    
\begin{table}[!ht]
\centering
\noindent
\begin{tabular}{P P P P G}\toprule
\textbf{Parameters} & \textbf{Hidden Size} & \textbf{Layers} & \textbf{Heads} & \textbf{Sequence Length} \\[0.1ex]
\midrule
2,795,443,200 & 2560 & 32 & 32 & 16384 \\ [1ex]
\midrule
\end{tabular}
\caption{Stable Code 3B model architecture.}
\label{tab:architecture-layout}
\begin{tabular}{H H H H}
\toprule
\textbf{Precision} & \textbf{Micro Batch Size} & \textbf{Gradient Accumulation Steps} & \textbf{Activation Checkpointing} \\[0.1ex]
\midrule
BF16 & 4 & 1 & enabled \\ [0.1ex]
\midrule
\end{tabular}
\caption{Stable Code 3B training configuration.}
\label{tab:training-config}
\end{table}
\section{Training}\label{sec:training-base}
In this section, we elaborate on the methodologies adopted for training the Stable Code models. This detailed analysis covers the compute infrastructure, setup configurations, and training techniques employed to optimize the model's learning process. We further delve into the use of the fill-in-the-middle (FIM) training objective and conduct a series of experiments to evaluate the effects of various initializations on the model's performance.

\subsection{Compute and Setup}
Stable Code was trained on 32 Amazon P4d instances comprising 256 NVIDIA A100 (40GB HBM2) GPUs. The size of our model, together with ZeRO stage 1 distributed optimization~\cite{rajbhandari2020zero}, eliminates the need for model sharding. Still, different triplets of micro-batch size, gradient accumulation steps, along with activation checkpointing granularity lead to different speed metrics. We employ a global batch size of $4,194,304$ tokens. With the setup in Table~\ref{tab:training-config}, we achieve $\approx$222 TFLOPs/s per device, or $71.15\%$ model flops utilization (MFU).


\subsection{Multi Stage Training}
\begin{figure}[!ht]
    \centering
    \includegraphics[width=\linewidth]{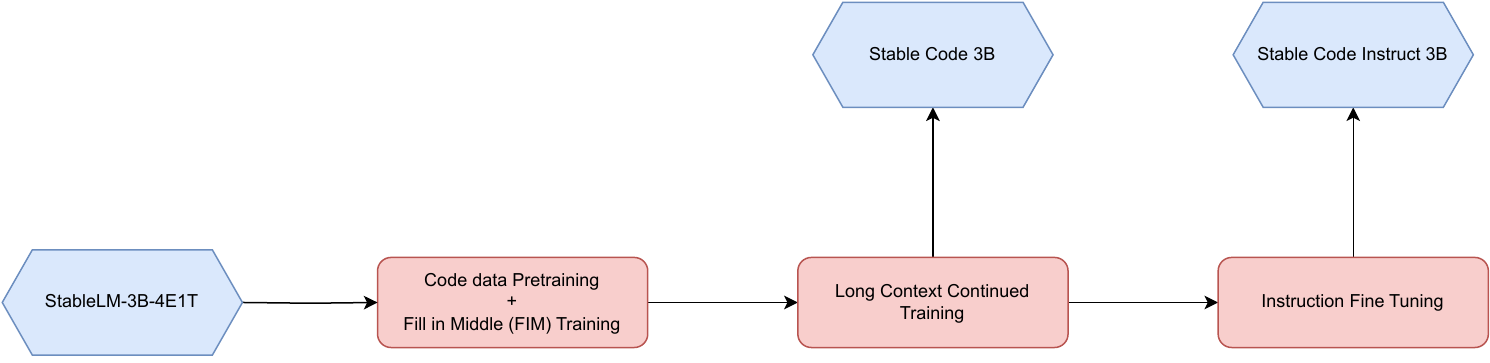}
    \caption{Staged approach to training Stable Code 3B and Stable Code Instruct 3B.}
    \label{fig:multistage}
\end{figure}

We employed a staged training approach that has been popular in other strong code language models such as the CodeGen \cite{Nijkamp2022CodeGenAO}, Stable Code Alpha \cite{StableCodeCompleteAlpha}, CodeLLaMA \cite{meta2023codellama}, and DeepSeekCoder \cite{guo2024deepseekcoder} models as shown in Figure~\ref{fig:multistage}.

We train Stable Code to predict the next token following standard autoregressive sequence modeling~\cite{radfordimproving}. We initialize our model from the Stable LM 3B checkpoint using the same base context length, $4096$ for the first stage of training with the data mix detailed in Section \ref{sec:pt-data}. This is followed by a continued pretraining stage as shown in Figure \ref{fig:multistage}.

\begin{figure}[!ht]
    \centering 
    \makebox[\textwidth]{\includegraphics[width=1.35\textwidth]{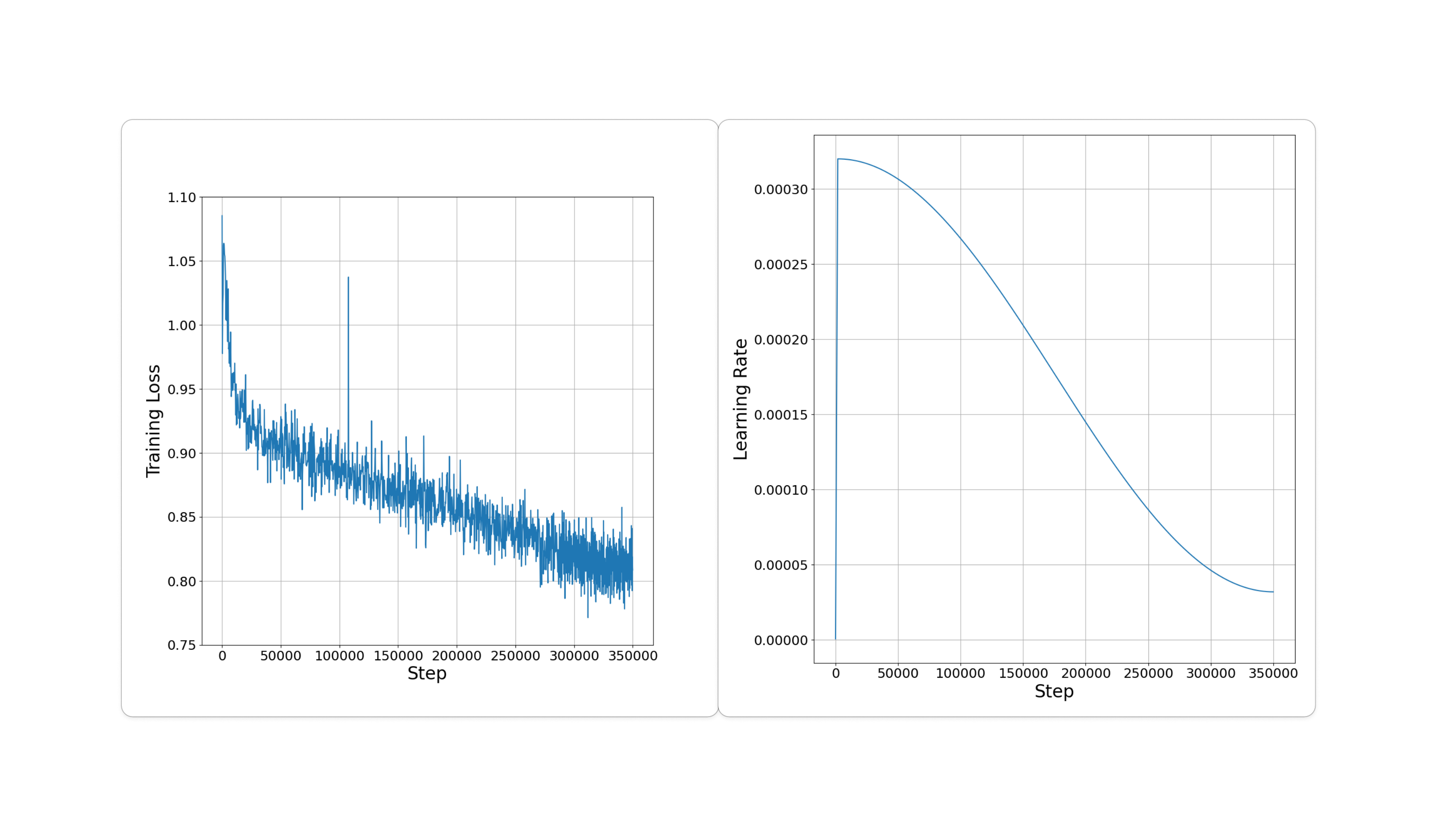}}
    \caption{Stable Code 3B Loss and Learning Rate Curves. 
    }
    \label{fig:loss-lr}
\end{figure}

Training is performed in BFloat16 mixed precision while keeping all-reduce operations in FP32. We use a standard AdamW optimizer with the following hyperparameters: $\beta_1 = 0.9, \beta_2 = 0.95, \epsilon = 1e-6,  \lambda (\text{weight decay}) = 0.1$. We start with a learning rate of 3.2e-4, set a minimum learning rate of 3.2e-5, and use a cosine decay learning rate schedule, shown in Fig \ref{fig:loss-lr}

The long context finetuning stage uses a peak learning rate of 2e-5 with a minimum of 1.28e-5 with the same betas, epsilon, and decay rate. We also switch the rotary embedding base to $1,000,000$ in line with CodeLLama~\cite{meta2023codellama} and Stable Code Alpha~\cite{StableCodeCompleteAlpha}. The long context finetuning stage is performed for another 20,000 steps, i.e.\ an additional $\approx$ 300 billion tokens.

\subsection{Model Initialization}
\begin{figure}[!ht]
    \makebox[\textwidth]{\includegraphics[width=1.3\textwidth]{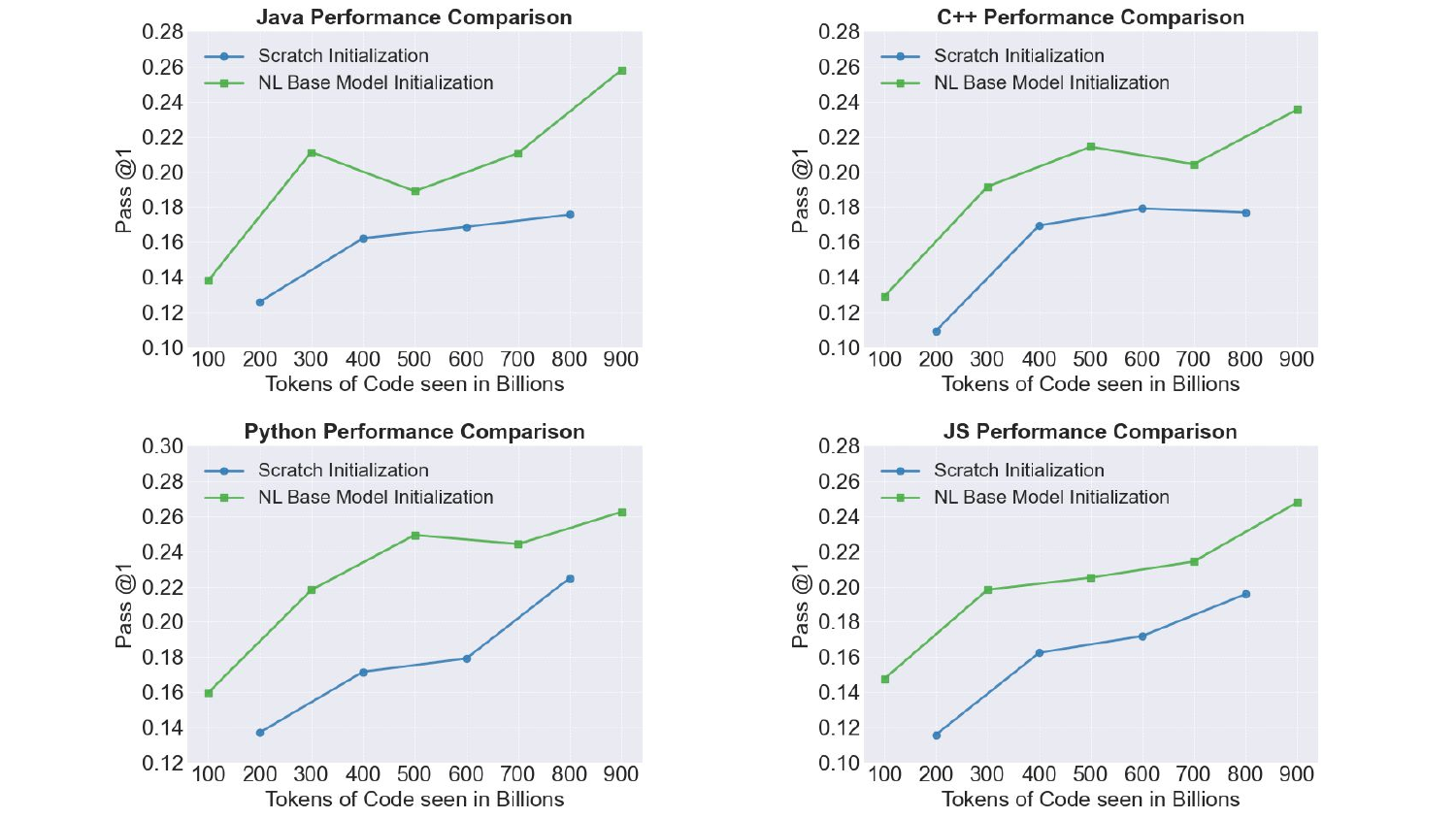}}
    \caption{Code Performance Comparison of Stable Code 3B Scratch and Stable LM 3B Initializations }
    \label{fig:stable_code_init}
\end{figure}

As with other types of models, code models mostly follow one of two main training recipes: models trained from scratch with code and related texts (e.g., CodeGen~\cite{Nijkamp2022CodeGenAO}, Stable Code Alpha~\cite{StableCodeCompleteAlpha}, Deepseek Coder~\cite{guo2024deepseekcoder}) and models leveraging continued pretraining from a foundational language model (e.g., works akin to CodeLLaMA~\cite{meta2023codellama}).
As possibly expected, we observe in Figure \ref{fig:stable_code_init} that models initialized from the weights of a pretrained language model (say \textit{Stable LM 3B} ~\cite{StableLM-3B-4E1T}) tend to outperform models trained from scratch.
This observation aligns with the hypothesis that a positive cross-transfer between natural language and code contributes to enhanced model capabilities, supported by insights from the "Naturalness of Software"~\cite{Devanbu2012OnTN} literature.


The Stable LM 3B base model was initially pre-trained on a dataset comprising 100 billion code tokens (in addition to 3.9 Trillion non-code tokens), which forms the baseline for all line charts depicted in green in Figure \ref{fig:stable_code_init}. This token count is consistently incorporated into the total token count for all subsequent training phases, ensuring an equi-code-token comparison across models. Figure \ref{fig:stable_code_init} shows us that at no point the line charts in blue cross the performance of those in green, suggesting that training on natural language tokens helps improve code capabilities of language models.


\subsection{Fill in the Middle (FIM) Training}
One of the core assumptions for natural language models training is the left-to-right causal ordering of tokens (note there are some exceptions, like Arabic).
In the case of code, this assumption does not always hold (e.g., function calls and function declarations can be in an arbitrary order for many programming languages). To address this, we use the "Fill in the Middle" objective \cite{Bavarian2022EfficientTO}. The approach involves splitting a document randomly into three segments: prefix, middle, and suffix, and then moving the middle segment to the end of the document. After the rearrangement the same autoregressive training process is followed.

This rearrangement allows for conditioning on structural patterns besides the prefix-only form used in traditional causal language modeling. The augmented data can then be further organized into two modes: ``Suffix-Prefix-Middle'' (``SPM'') and ``Prefix-Suffix-Middle'' (``PSM''). Following \cite{Bavarian2022EfficientTO}, we apply FIM at the character-level using a rate of 50\% and choose between ``SPM'' and ``PSM'' modes with uniform probability.

FIM was applied during both stages of pretraining. In order to account for FIM in the long context training phase, we made sure to only allow FIM to be applied within the confines of individual files to avoid introducing unrealistic scenarios into the training objective.


\section{Fine-tuning and Alignment} \label{sec:alignment}

Following pre-training, we further improve our model's conversational skills via a fine-tuning stage that consists of supervised fine-tuning (SFT) and direct preference optimization (DPO)~\cite{rafailov2023direct}.

We use the following datasets that are publicly available on Hugging Face for SFT finetuning: OpenHermes \footnote{https://huggingface.co/datasets/teknium/OpenHermes-2.5}, Code Feedback \footnote{https://huggingface.co/datasets/m-a-p/CodeFeedback-Filtered-Instruction}, CodeAlpaca \footnote{https://huggingface.co/datasets/HuggingFaceH4/CodeAlpaca\_20K}. Together they provide around $500,000$ training samples after performing an exact match deduplication.
We train our SFT models for three epochs using a cosine learning rate scheduler. A warm-up phase of $10\%$ of the training duration is employed before reaching the peak learning rate of 5e-5. We set the global batch size to 512 sequences and pack inputs into sequences of up to 4096 tokens in length.

After SFT, we apply Direct Preference Optimization (DPO)~\cite{radfordimproving}, a technique that has been key to the success of recent high-performing models like Zephyr-7B \cite{tunstall2023zephyr}, Neural-Chat-7B, and Tulu-2-DPO-70B \cite{ivison2023camels}, Stable LM 2 1.6B Zephyr \cite{bellagente2024stable}.
In this case, we curated a dataset of approximately 7,000 samples, leveraging data from  UltraFeedback \cite{cui2023ultrafeedback}, Distilabel Capybara DPO-7k Binarized \footnote{https://huggingface.co/datasets/argilla/distilabel-capybara-dpo-7k-binarized} and only kept samples related to code by filtering using an LLM based approach. Additionally, to improve the model safety we also included the Helpful and Harmless RLFH dataset from Bai et al.\cite{bai2022training}

Moreover, we added the harmless subset from the HH-Anthropic dataset \footnote{https://github.com/anthropics/hh-rlhf/}. We curated a dataset comprising instances that received safety ratings of 0 or 1, indicating content that is critical for training models to avoid generating or perpetuating harmful outputs. This process resulted in the compilation of approximately 15,000 high-relevance safety-related data points.

We incorporate a DPO step that adheres to the Zephyr recipe, as outlined by Tunstall et al \cite{tunstall2023zephyr}. The tuning of the $\beta$ hyperparameter, which is set to $0.01$, following best practices for ensuring model stability and convergence. Furthermore, we employ RMSProp as our optimization algorithm, the initial phase of DPO training involves increasing the learning rate to a peak value of 5e-7, with $10\%$ warm-up step of the total training duration.



\section{Results}\label{sec:results}
Throughout this section, we showcase a comparative analysis of the Stable Code and Stable Code Instruct models, with a range of selected baselines and models previously recognized as state-of-the-art within the same computational scale of 3B parameters. We also look at larger code language models, illustrating the diverse competitive landscape our model navigates.

\subsection{Code Completion Benchmarks}
\subsubsection{Stable Code Base}
The primary metric for this comparison is the model's performance on code completion tasks, a fundamental evaluation criterion given its direct relevance to the practical utility of code language models. We use the Multi-PL benchmark proposed by Cassano et al., \cite{multipl} to evaluate the models.

\begin{xltabular}{\textwidth}{l|r|c| *{6}{c} r}
\caption{Performance of different base code LLMs with size 3B params and under on Multi-PL.} \label{tab:performance} \\
\toprule
Model & Size & Avg & \rotatebox{0}{Python} & \rotatebox{0}{C++} & \rotatebox{0}{JavaScript} & \rotatebox{0}{Java} & \rotatebox{0}{PHP} & \rotatebox{0}{Rust} \\
\midrule
\endfirsthead
\toprule
\endhead

\bottomrule
\endlastfoot
Stable Code & 3B & 29.1 & \textbf{32.4} & 30.9 &  32.1 & \textbf{32.1 }& 24.2 & 23.0\\
StarCoder v2 & 3B &\textbf{ 30.9}& 28.9 & \textbf{32.3} & \textbf{36.5} & 31.2 & \textbf{29.9} & \textbf{25.6} \\
DeepSeek Coder & 1.3B & 26.1 & 28.6 & 29.2 & 28.7 & 29.0 & 23.6 & 18.5\\
Wizard Coder & 3B &25.7 & 31.2 & 25.6 & 26.2 & 25.8 & 25.3 & 20.4\\
Replit Code V1.5 & 3B & 23.9& 23.0 & 25.9 & 26.2 & 23.6 & 23.2 & 21.5\\
StarCoder & 3B &19.9& 21.6 & 19.8 & 21.5 & 20.5 & 19.0 & 16.9\\
Deci Coder & 1B & 10.8& 19.1 & 6.8 & 18.4 & 16.7 & 2.1 & 1.7\\
\midrule
Deepseek Coder & 6.7B &  \textbf{43.0} & \textbf{45.9} & \textbf{48.7} & \textbf{46.1} & \textbf{43.0} & \textbf{38.6} & \textbf{35.4}\\
Code Llama & 7B & 29.1& 30.0 & 28.2 & 32.5 & 31.9 & 25.7 & 26.3\\
StarCoder & 15B & 29.0& 33.6  & 31.6 & 30.8 & 30.2 & 26.1 & 21.8\\
\end{xltabular}

Despite its relatively smaller size --less than 40\% and 20\% the number of parameters of Code Llama~\cite{meta2023codellama} and StarCoder 15B~\cite{li:starcoder}, respectively-- Stable Code matches their performance on average across programming languages.
Also, Stable Code 3B achieves strong performance at the 3B scale, showing remarkable capabilities in code completion tasks.
The StarCoder v2 model~\cite{lozhkov2024starcoder} --which is a newer model trained on significantly more data-- outperforms Stable Code 3B base on average.
\footnote{See the detailed and comprehensive work by the BigCode team \cite{lozhkov2024starcoder} for updated results around that model.}

\subsubsection{Stable Code Instruct}
Similarly, we evaluate instruct-tuned variants of these models on the Multi-PL benchmark with Stable Code Instruct in Table \ref{tab:instruct-multipl}.




\begin{xltabular}{\textwidth}{G M M Q Q Q Q Q Q}
\caption{Performance of instruction tuned code LLMs with size around 3B params on Multi-PL.} \label{tab:instruct-multipl} \\
\toprule
\multicolumn{1}{l}{Model} & Size & Avg & \rotatebox{0}{Python} & \rotatebox{0}{C++} & \rotatebox{0}{JS} & \rotatebox{0}{Java} & \rotatebox{0}{PHP} & \rotatebox{0}{Rust} \\
\midrule
\endfirsthead
\toprule
\endhead

\bottomrule
\\
\endfoot

\bottomrule
\endlastfoot

\multicolumn{1}{l}{Stable Code Instruct} & 3B & \textbf{47.2} & \textbf{58.6} &\textbf{ 48.1} & 49.2 & \textbf{44.4} & 45.6 & \textbf{37.2} \\
\multicolumn{1}{l}{DeepSeek Coder Ins} & 1.3B & 44.3 & 52.5 & 45.0 &\textbf{ 52.3 }& 40.9 & \textbf{46.4} & 28.6 \\
\midrule
\multicolumn{1}{l}{DeepSeek Coder Ins} & 6.7B & \textbf{61.1} & \textbf{64.6} & \textbf{63.2 } &\textbf{ 67.7 } & \textbf{59.1 } & \textbf{62.7 } & \textbf{48.9 } \\
\multicolumn{1}{l}{Codellama Instruct}& 7B & 30.6 & 32.7 & 30.8 & 33.6 & 31.5 & 29.6 & 25.5 \\
\end{xltabular}
Stable Code Instruct offers a remarkably solid performance for its size.





\subsection{Fill in the Middle}
The capacity to utilize both preceding and succeeding context significantly benefits non-instruct tuned code language models, enabling more precise and context-aware completions.
Therefore, we further extend our evaluation to the specialized Fill in the Middle (FIM) code completion performance. In this case, we utilize the StarCoder-FIM/SantaCoder-FIM evaluation tasks from the BigCode Evaluation Harness~\cite{bigcode-evaluation-harness}, offering a nuanced assessment of each model's understanding and prediction capabilities in more complex coding contexts.
We focus on the best performing models from Table \ref{tab:performance}, and Table \ref{tab:fim-performance} shows the FIM performance of such models.

\begin{xltabular}{\textwidth}{l|r| *{3}{c} r}
\caption{FIM performance of different base code models with size at most 3B parameters.} \label{tab:fim-performance} \\
\toprule
Model & Size & \rotatebox{0}{Python} & \rotatebox{0}{JavaScript} & \rotatebox{0}{Java} \\
\midrule
\endfirsthead
\toprule
\endhead

\bottomrule
\endlastfoot
Stable Code & 3B & \textbf{59.1} & \textbf{73.4 }& 64.1 \\
StarCoder V2 & 3B & 59.0 & 72.8 & \textbf{74.9}\\
DeepSeek Coder & 1.3B & 53.5 & 65.2 & 49.4\\
StarCoder & 3B & 50.3 & 61.4 & 56.1\\
\end{xltabular}

\subsection{Multi Turn Benchmark}
We also evaluate instruct tuned models on the code subset of the challenging Multi-turn benchmark (MT-Bench) ~\cite{zheng2023judging}.
Table \ref{tab:instruct-mtb} shows the results of coding questions in MT-Bench.
\begin{xltabular}{\textwidth}{l|r|*{1}{c} r}
\caption{MT-Bench Coding Question Score.} \label{tab:instruct-mtb} \\
\toprule
Model & Size & \rotatebox{0}{MT-Bench[Coding]}\\
\midrule
\endfirsthead
\toprule
\endhead
Stable Code Instruct & 3B & \textbf{5.8 }\\
DeepSeek Coder & 1.3B & 4.6 \\
\midrule
DeepSeek Coder & 6.7B & \textbf{6.9} \\
CodeLlama Instruct & 7B & 3.6\\ 
\midrule

\end{xltabular}

\subsection{SQL Performance}
One important application for code language models are database query tasks. In this domain, we compare the performance of Stable Code Instruct against other popular instruction-tuned models and models specifically trained to perform well in SQL.
We use the benchmark created by Defog AI \footnote{https://defog.ai/blog/open-sourcing-sqleval/} to evaluate our models.
\begin{xltabular}{\textwidth}{l|r|c|c *{5}{c} r} 
\caption{Evaluation of popular models and Stable Code Instruct 3B on SQL-Eval.} \label{tab:sql-eval-transposed} \\
\toprule
 & Size & Avg & Date & Group By & Order By & Ratio & Join & Where \\
\midrule
\endfirsthead
\toprule
\endhead

\bottomrule
\endfoot

\bottomrule
\endlastfoot

Stable Code Instruct & 3B & \textbf{47.2} &24.0 & 54.2 & 68.5 & 40.0 & 54.2 & 42.8 \\
DeepSeek-Coder Instruct & 1.3B & 34.2 & 24.0 & 37.1 & 51.4 & 34.3 & 45.7 & 45.7 \\
\midrule
SQLCoder & 7B & \textbf{70.6}  & 64.0 & 82.9 & 74.3 & 54.3 & 74.3 & 74.3 \\
\end{xltabular}







\section{Inference} \label{sec:inference}

Speed and memory considerations are particularly important for code models.
Accordingly, in this section we briefly mention how to make Stable Code even faster.
In our release, we offer quantized weights for the Stable Code model \footnote{\url{https://huggingface.co/stabilityai/stable-code-3b/tree/main}}, ensuring compatibility with widely adopted inference libraries such as llama.cpp \footnote{\url{https://github.com/ggerganov/llama.cpp}} and Apple
MLX~\cite{mlx2023}.

\subsection{Quantization}

We provide quantized model files in various formats to facilitate seamless integration with a diverse range of inference engines. The offered quantization files encompass the following models: GGUF Q5\_K\_M, GGUF Q6\_K, GGUF FP16, and MLX INT4.

\subsection{Throughput}
In Table \ref{tab:quant-tput}, we present the throughput numbers obtained when running Stable Code on consumer-grade devices and the corresponding system environments. These results indicate a nearly two-fold increase
in throughput when employing lower precision. However, it is important to note that these data points serve as a rough estimate and do not represent the outcome of comprehensive benchmarking. Instead, they aim to offer users with a
practical understanding of potential performance improvements on typical hardware.

It is worth mentioning that implementing lower precision quantization may lead to some reduction in model performance (maybe significant). Therefore, we strongly advise researchers and developers to evaluate the actual impact on their specific
use cases in real-world scenarios before making decisions.

\begin{xltabular}{\textwidth}{P | P | N | P | P}
\caption{Throughput and power usage on various devices using different quantization frameworks.} \label{tab:quant-tput} \\
\toprule
Framework & CPU & Precision & Throughput (Tok/s) & Power Consumption (W) \\
\midrule
\endfirsthead
\toprule
\endhead

\bottomrule
\endlastfoot

MLX & M2 Pro Max & FP16 & 23 & 18 \\
MLX & M2 Pro Max & INT4 & 52 & 17 \\
GGUF & M2 Pro Max & FP16 & 28 & 14 \\
GGUF & M2 Pro Max & Q5\_K\_M & 53 & 23 \\
GGUF & M2 Pro Max & Q6\_K & 54 & 23 \\
\end{xltabular}

\section{Conclusion} \label{sec:conclusion}
In this report, we introduce Stable Code and Stable Code Instruct, two compact decoder-only language models targeting different software development use cases.
In the spirit of open science, we detail all datasets for pre-training and alignment, and discuss our training and evaluation methodologies. We also conduct extensive model evaluations and comparisons with other similarly-sized models, demonstrating Stable Code and Stable Code Instruct's remarkable performance.
Finally, we profile the model on common edge computing architectures.
We hope this report and the models we release contribute to the improvement and further research on code language models and their use.

{\small
\bibliographystyle{plain}
\bibliography{literature}
}
\newpage
%
%
\appendix
\section{Examples} \label{sec:examples}
In this appendix, we briefly showcase some examples of Stable Code in action. The context is shown in black whereas the suggestion completion is shown in magenta.

\definecolor{codegreen}{rgb}{0,0.6,0}
\definecolor{codegray}{rgb}{0.5,0.5,0.5}
\definecolor{codepurple}{rgb}{0.58,0,0.82}
\definecolor{backcolour}{rgb}{0.95,0.95,0.92}

\lstdefinestyle{mystyle}{
  backgroundcolor=\color{backcolour}, 
  commentstyle=\color{magenta},
  keywordstyle=\color{black},
  numberstyle=\tiny\color{codegray},
  stringstyle=\color{black},
  basicstyle=\ttfamily\footnotesize,
  breakatwhitespace=false,         
  breaklines=true,                 
  captionpos=b,                    
  keepspaces=true,                 
  numbers=left,                    
  numbersep=5pt,                  
  showspaces=false,                
  showstringspaces=false,
  showtabs=false,                  
  tabsize=2,
  escapeinside={(*@}{@*)}, 
}

\lstset{style=mystyle}
\begin{lstlisting}[language=Python, caption=Bubble Sort FIM Example.]
def bubbleSort(arr):
    n = len(arr)

    for i in range(n):
        swapped = False

        for j in range(0, n-i-1):
            # if arr[j] > arr[j+1]:
                arr[j], arr[j+1] = arr[j+1], arr[j]
                swapped = True
        
        if swapped == False:
            break

\end{lstlisting}

\lstset{style=mystyle}
\begin{lstlisting}[language=Python, caption=Typescript Completion Example.]
        return chat_model.invoke(messages);
    }
}

export function fetchResponse(prompt: string, fileName: string, MODEL_NAME: string, API_KEY: string, modelType: string): Promise<AIMessage> | Promise<string> {
  console.log("Fetching response", modelType);
  switch (modelType) {
    case "action":
      return createChatModel(MODEL_NAME, API_KEY).invoke([new HumanMessage(prompt)]);
    case "completion":
      return createCompletionModel(MODEL_NAME, API_KEY).invoke(prompt);
    default:
      throw new Error(#`Unknown model type: ${modelType}`);
  }
}
\end{lstlisting}

\lstset{style=mystyle}
\begin{lstlisting}[language=Python, caption=PyTorch Modeling Example.]
stage_3 = DiffusionPipeline.from_pretrained(
    "stabilityai/stable-diffusion-x4-upscaler", **safety_modules, torch_dtype=torch.float16
)

stage_3.enable_model_cpu_offload()

prompt = 'a cute otter'
generator = torch.manual_seed(1)

prompt_embeds, negative_embeds = stage_1.encode_prompt(prompt)

stage_1_output = stage_1(
    prompt_embeds=#prompt_embeds, negative_embeds=negative_embeds, generator=generator
).images

stage_2_output = stage_2(
    image=stage_1_output,
    prompt_embeds=prompt_embeds,
    negative_prompt_embeds=negative_embeds,
    generator=generator,
    output_type="pt",
).images

stage_3_output = stage_3(prompt=prompt, image=stage_2_output, noise_level=100, generator=generator).images
make_image_grid([pt_to_pil(stage_1_output)[0], pt_to_pil(stage_2_output)[0], stage_3_output[0]], rows=1, rows=3)

\end{lstlisting}
\end{document}